%% file: main.tex
\documentclass[10pt,twocolumn,letterpaper]{article}

 \usepackage{iccv}              


\definecolor{iccvblue}{rgb}{0.21,0.49,0.74}
\usepackage[pagebackref,breaklinks,colorlinks,allcolors=iccvblue]{hyperref}
\usepackage{algorithm}
\usepackage{algorithmicx}
\usepackage{algcompatible} 

\usepackage{booktabs}
\usepackage{multirow}
\usepackage{amsmath}
\usepackage{amssymb}
\usepackage{bbding} 
\usepackage{pifont}
\usepackage{tcolorbox}

\usepackage{listings}
\makeatletter  
\newif\if@restonecol  
\makeatother  

\usepackage{algorithm}
\usepackage{algorithmicx}
\usepackage{algpseudocode}

\def\paperID{7420} 
\def\confName{ICCV}
\def\confYear{2025}

\title{Open-Vocabulary Octree-Graph for 3D Scene Understanding}

\author{Zhigang Wang$^{1,2}$\thanks{Equal contribution.}, Yifei Su$^{3,4,2*}$, Chenhui Li$^{2*}$,\\ Dong Wang$^{2}$, Yan Huang$^{3,4}$, Xuelong Li$^{5}$, Bin Zhao$^{1,2}$\thanks{Corresponding author.}\\
$^{1}$Northwestern Polytechnical University,
$^{2}$Shanghai AI Laboratory,\\
$^{3}$University of Chinese Academy of Sciences,
$^{4}$CASIA,
$^{5}$TeleAI
}

\begin{document}
\maketitle

\input{sec/0_abstract}    
\input{sec/1_intro}
\input{sec/2_related}
\input{sec/3_method}

\input{sec/4_experiment}

\input{sec/5_conclusion}

\section*{Acknowledgments}
This work is supported by the Shanghai AI Laboratory, the National Natural Science Foundation of China (62376222), and Young Elite Scientists Sponsorship Program by CAST (2023QNRC001).

{
    \small
    \bibliographystyle{ieeenat_fullname}
    \bibliography{main}
}

\end{document}

%% file: sec/0_abstract.tex
\begin{abstract}

Open-vocabulary 3D scene understanding is indispensable for embodied agents. Recent works leverage pretrained vision-language models (VLMs) for object segmentation and project them to point clouds to build 3D maps. Despite progress, a point cloud is a set of unordered coordinates that requires substantial storage space and does not directly convey occupancy information or spatial relation, making existing methods inefficient for downstream tasks, e.g., path planning and text-based object retrieval. To address these issues, we propose \textbf{Octree-Graph}, a novel scene representation for open-vocabulary 3D scene understanding. Specifically, a Chronological Group-wise Segment Merging (CGSM) strategy and an Instance Feature Aggregation (IFA) algorithm are first designed to get 3D instances and corresponding semantic features. Subsequently, an adaptive-octree structure is developed that stores semantics and depicts the occupancy of an object adjustably according to its shape. Finally, the Octree-Graph is constructed where each adaptive-octree acts as a graph node, and edges describe the spatial relations among nodes. Extensive experiments on various tasks are conducted on several widely-used datasets, demonstrating the versatility and effectiveness of our method.
Code is available \href{https://github.com/yifeisu/OV-Octree-Graph}{here}.
\end{abstract}

%% file: sec/1_intro.tex
\section{Introduction}
\label{sec:intro}

3D scene understanding is receiving increasing attention due to its widespread usage in robots \cite{3DOGNav} and VR/AR applications \cite{lerf}. Previous works \cite{mask3d, MVA, isbnet, ODAM, vmap} trained models on particular 3D scene datasets to complete this task. Although significant progress has been achieved, they are limited to a closed-set category. Recently, we have witnessed the impressive generalization ability of foundation models (\eg, SAM \cite{SAM} and CLIP \cite{CLIP}) which can perceive various objects in unseen scenarios, inspiring a lot of open-vocabulary 3D scene understanding methods \cite{ovir3d, conceptgraph, hovsg, 3DOVS, Lowis3D, PLA, RegionPLC}. Given an RGB-D sequence with camera poses, mainstream methods leverage the off-the-shelf foundation models to generate 2D object masks and corresponding visual-language features, and then project them to point clouds to construct a semantic 3D map.


\begin{figure}[t]
    \centering
    \includegraphics[width=0.4\textwidth]{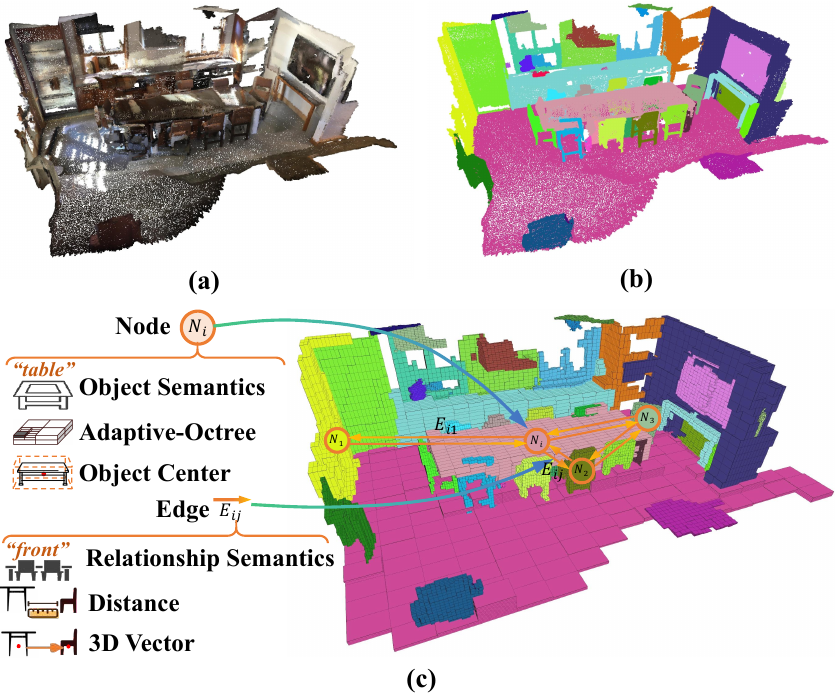} 
    \caption{(a) A 3D scene. (b) The corresponding semantic 3D map based on point clouds (6.8M). (c) Our Octree-Graph where each object is represented by the proposed adaptive-octree and each edge contains rich spatial relations among objects. All adaptive-octrees occupy 42KB of storage space in total.}
    \label{Figure:intro}
    \vspace{-2mm}
\end{figure}

Despite the favorable open-vocabulary understanding capability, they have two drawbacks. 1) \textbf{\textit{Inefficient space representation of 3D scenes}}. Most mainstream methods \cite{OpenScene, maskclustering, conceptfusion} build the 3D map based on point clouds, as shown in Fig. \ref{Figure:intro} (b). Point clouds are unordered discrete coordinates that require considerable storage space, making existing methods inefficient to deploy on embodied agents with limited storage resources. Moreover, point clouds lack explicit representation of occupancy information and spatial connectivity which are critical for downstream tasks, \eg, path planning and text-based object retrieval. 2) \textbf{\textit{Inaccurate semantic object segmentation for 3D map construction}}. Most methods overlook the inaccuracy of foundation models/vision-language models (VLMs) when conducting object segmentation and feature extraction, inevitably causing imprecise 3D object segments and degraded semantics.

To alleviate these problems, we propose \textbf{Octree-Graph} as shown in Fig. \ref{Figure:intro} (c), a novel open-vocabulary scene representation designed to characterize the occupancy and semantics of each object, as well as the relations among them. Specifically, the adaptive-octree is first proposed to depict each object's occupancy, which inherits the advantages of the octree structure by hierarchically representing a 3D space with structured sub-regions. Compared to the point cloud without regional or hierarchical information, it can save significant storage space. Furthermore, our adaptive-octree initializes each object adaptively according to its shape, enabling a precise description of the occupancy within a limited octree depth. This is particularly suitable for objects with large aspect ratios, \eg, walls and floors. Based on this, the Octree-Graph is constructed where each adaptive-octree acts as a graph node, and each edge encompasses rich relations among objects, \eg, distances and relative orientations. The proposed Octree-Graph can be directly applied to downstream tasks such as object retrieval, occupancy queries, and path planning, thus providing significant convenience.

To obtain accurate semantic objects for Octree-Graph construction, we devise a \textbf{training-free} pipeline. First, given input images, 2D proposals are segmented via an off-the-shelf segmenter, and corresponding visual-language features are extracted by pretrained VLMs. Then, they are projected into 3D space as point cloud segments. Second, to correctly merge segments belonging to the same instance, a Chronological Group-wise Segment Merging (CGSM) strategy is proposed, where the segments are partitioned into several groups in time order. Each group is individually processed to leverage spatiotemporal details from the neighborhood while avoiding interference from global redundancy. 
Third, an Instance Feature Aggregation (IFA) method is proposed to obtain semantic representations for each object. Unlike existing works that directly average features as a result, we simultaneously consider the representativeness and distinctiveness of a feature during the fusion process. Our contributions are summarized as follows.

\begin{itemize}
    \item We propose the Octree-Graph for open-vocabulary 3D scene understanding, which efficiently depicts objects' occupancies, semantics, and relations, benefiting several downstream tasks.
    
    \item We propose a Chronological Group-wise Segment Merging (CGSM) strategy and an Instance Feature Aggregation (IFA) method to obtain accurate semantic objects.

    \item We conduct extensive experiments, demonstrating the versatility, effectiveness, and efficiency of our method. 
\end{itemize}


%% file: sec/2_related.tex
\section{Related Work}
\label{sec:related}

\textbf{Foundation Models.} Recently, foundation models have exhibited impressive zero-shot perception ability. Here, we review several foundation models related to our work. CLIP \cite{CLIP} is a popular vision-language model that associates images and texts through contrastive learning, significantly promoting many vision-language tasks. SAM \cite{SAM} is a class-agnostic 2D segmentation model trained with over 1 billion masks, demonstrating powerful zero-shot performance. OVSeg \cite{OVSeg} finetunes CLIP to gain the ability of open-vocabulary semantic segmentation. CropFormer \cite{Cropformer} fuses the full image and high-resolution image crops to improve segmentation performance. TAP \cite{Tap} can simultaneously conduct recognition, segmentation, and caption generation. Additionally, many other methods \cite{LSeg, openseg, groupvit, ovod, simpleovod, grounding-dino, LG3DIS, SPD, OVIS, MFOVIS} are proposed for 2D open-vocabulary object detection and segmentation.


\begin{figure*}[t]
    \centering
    \includegraphics[width=0.975\textwidth]{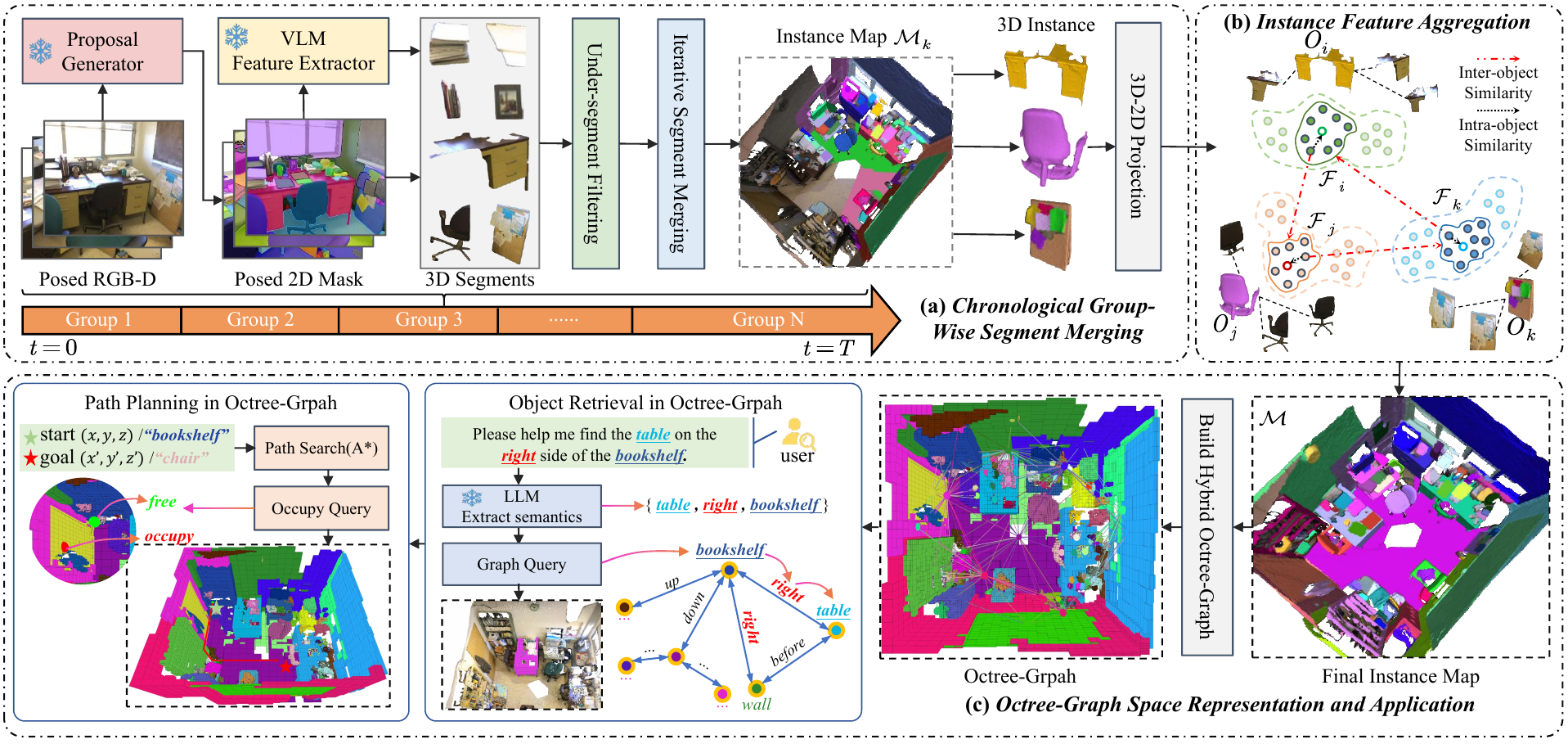} 
    \caption{Overview of our Octree-Graph. (a) Chronological Group-wise Segment Merging (CGSM). Given posed RGB-D inputs, 2D masks with semantic features are first extracted and then projected into the 3D space, where CGSM is conducted to merge segments. (b) Instance Feature Aggregation (IFA). Feature aggregation is performed for each merged object, which considers both intra- and inter-object similarity. (c) The Octree-Graph is constructed to efficiently and accurately represent the scene, facilitating various downstream tasks.}
    \label{fig:overview}
    \vspace{-2mm}
\end{figure*}

\noindent \textbf{Open-Vocabulary 3D Scene Understanding.} Based on the organization form of scene representation, we categorize these works into four types. 1) \textbf{\textit{NeRF/Gaussian 3D mapping}}. These methods perform 3D scene understanding and scene/object reconstruction simultaneously, \emph{e.g.}, OpenObj \cite{openobj}. Although achieving good performance, they need extra effort to train the NeRF or 3D Gaussian models. 2) \textbf{\textit{point/grid-wise 3D mapping}}. This branch involves directly projecting semantic features to each 3D point. OpenScene \cite{OpenScene} and ConceptFusion \cite{conceptfusion} extract CLIP features from the images and densely project them to the point cloud. VLMaps \cite{VLMaps} adopts a similar pipeline to project visual-language features to a grid-based BEV map. 3) \textbf{\textit{instance-wise 3D mapping}}. These works explicitly obtain each 3D instance and fuse its visual-language features for 3D mapping. OpenIns3D \cite{OpenIns3D} is a 3D-input-only framework that gets objects by open-vocabulary 3D detection. OVIR-3D \cite{ovir3d}, SAM3D \cite{sam3d}, and MaskClustering \cite{maskclustering} follow a 2D-to-3D pipeline where 2D masks are projected into 3D space for instance merging based on semantic similarity and 3D overlap. SAI3D \cite{sai3d} uses both 2D proposals and 3D super points for instance segmentation. OpenMask3D \cite{openmask3d}, Open3DIS \cite{open3dis}, and SA3DIP \cite{SA3DIP} leverage extra 3D instance detectors to get more accurate object proposals. However, the used 3D models cannot be considered purely zero-shot methods. 4) \textbf{\textit{3D scene graph/octree}}. A few works use a graph or octree to organize the scene. ConceptGraph \cite{conceptgraph} and Clio \cite{clio} cluster object segments and construct a scene graph to enhance spatial reasoning. HOV-SG \cite{hovsg} proposes a hierarchical 3D scene graph to enable scene representation of different granularities. OctreeOcc \cite{octreeocc} and PlenOctrees \cite{plenoctree} use the octree structure to store semantic class and rendering information, respectively. In contrast, our Octree-Graph represents each object using an adaptive-octree and models their relations using a graph, supporting efficient occupancy queries and spatial reasoning.

%% file: sec/3_method.tex
\section{Method}
\label{sec:method}

\subsection{Framework Overview}
As shown in Fig \ref{fig:overview}, given a sequence of RGB images \(\mathcal{I}_c=\{\mathbf{I}_t^c\}_{t=1}^{T}\) and depth images \(\mathcal{I}_d=\{\mathbf{I}_t^d\}_{t=1}^{T}\) scanned in a scene, we first leverage VLMs to extract segment proposals (\S~\ref{sec:mask_generate}). Next, we chronologically merge these segments into an instance map \(\mathcal{M}\) via a group-wise merging strategy (\S~\ref{sec:merging}). Then we dynamically aggregate the redundant semantics of each instance into a distinctive feature (\S~\ref{sec:feature_aggregate}). Finally, we build an Octree-Graph \(G\) to represent spatial relations among instances, with the adaptive-octree to detail instance occupancy. Based on this, we implemented LLM-based object retrieval and path planning algorithms (\S~\ref{sec:og_application}). 

\subsection{Segment Proposal and Comprehension} \label{sec:mask_generate}
For each frame \(\mathbf{I}^c_t\) at time $t$, we first adopt an off-the-shelf proposal generator, \eg, CropFormer \cite{Cropformer}, to extract a set of 2D masks \(\mathcal{P}_t^{2d}=\{ \mathbf{m}_{i} \}_{i=1}^{n_t}\), where \(n_t\) is the mask number. We then filter out tiny and marginal masks to ensure the proposal quality. 
Next, each \(\mathbf{m}_{i}\) is fed into the visual encoder and caption generator to obtain the visual feature \(\mathbf{f}_{i}^{v}\) and caption feature \(\mathbf{f}_{i}^{c}\).
Finally, we project each mask \(\mathbf{m}_{i}\) into the 3D space as a point cloud segment and perform DBSCAN \cite{DBSCAN} denoise, obtaining segments \(\mathcal{P}_t^{3d}=\{ \mathcal{S} _{i} \}_{i=1}^{n_{t}}\).

\subsection{Chronological Group-wise Segment Merging} \label{sec:merging}
Existing segment merging strategies are typically categorized into two types: 1) \textit{frame-wise}, which sequentially or hierarchically merges the adjacent frames \cite{conceptgraph, ovir3d}, integrating similar segments efficiently. 2) \textit{graph-wise}, which merges segments across all frames \cite{hovsg, maskclustering}.
These methods have achieved great success, while the former solely relying on a single frame can be easily affected by proposal noises, \eg, associating unrelated instances once an under-segment is merged.
The latter, which processes all segments together, may introduce redundant computations and be affected by irrelevant segments.
To this end, we propose a Chronological Group-wise Segment Merging (CGSM) strategy with semantic-guided under-segment filtering and a dynamic threshold decay strategy.

\begin{figure}[t]
    \centering
    \includegraphics[width=0.45\textwidth]{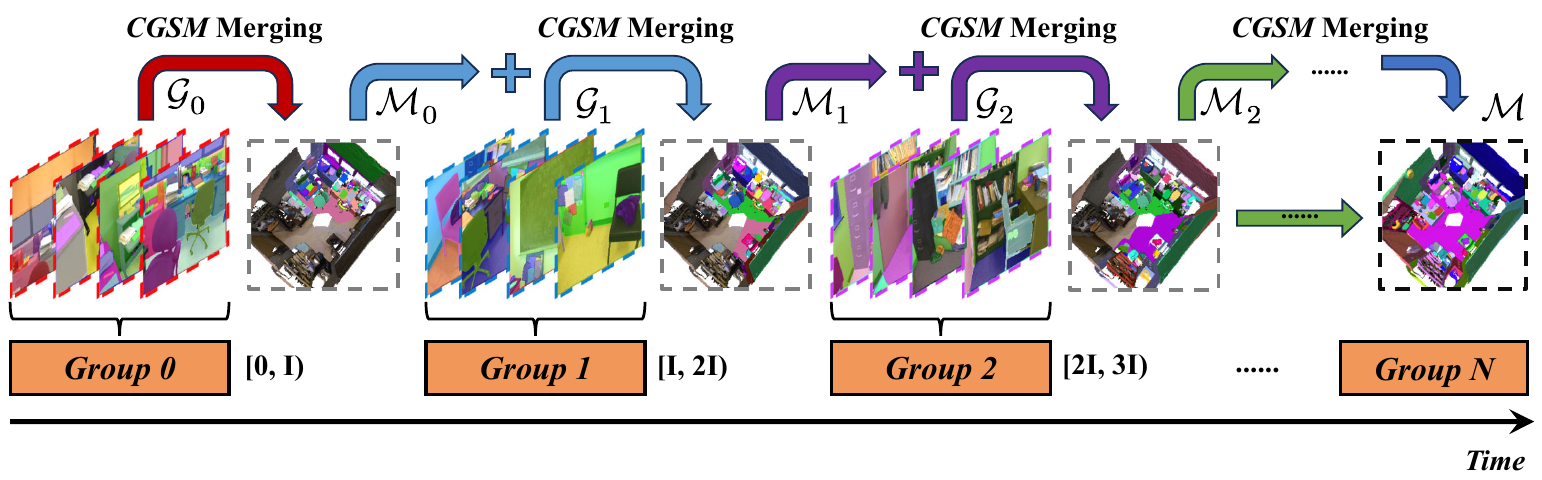} 
    \caption{Illustration of group split and CGSM merging.}
    \label{Figure:CGSM}
    \vspace{-4mm}
\end{figure}

\noindent\textbf{Chronological Group-Wise Split.}
Given the prior that \textit{an instance often appears in multiple consecutive frames}, as shown in Fig. \ref{Figure:CGSM}, CGSM first partitions all frames into several groups in time order with interval \(I\), obtaining the set of segments \( \mathcal{G}_i \) for each group. In this way, a group can retain the adjacent segments while avoiding interference caused by global segments. 
Based on these groups, we perform iterations of merging to integrate separate segments into an instance map \(\mathcal{M}\). 
Concretely, we start by merging \(\mathcal{G}_0\) into an intermediate instance map \(\mathcal{M}_0\).
Subsequently, we iteratively take the union \(\{\mathcal{M}_{k-1}, \mathcal{G}_k\}\) as input for the \(k^{\mathrm{th}}\) merging, until the final instance map \(\mathcal{M}\) is constructed. Next, we elaborate on the details of a single merging step.

\noindent\textbf{Segment Group Merging.} 
For two segments \(\{\mathcal{S}_m, \mathcal{S}_n\}\), we define \(\phi_{\mathrm{sem}}^{{v}}(m,n)\) as the cosine similarity between their visual features, and \(\phi_{\mathrm{sem}}^{{c}}(m,n)\) the cosine similarity between caption features.
Regarding geometric similarity, we compute \(\phi_{\mathrm{geo}}^{\mathrm{iou}}(m,n)\) as the intersection over union of two segments. Additionally, we calculate the ratio of \(\mathcal{S}_n\) contained within \(\mathcal{S}_m\) as \(\phi_{\mathrm{geo}}^{\mathrm{ior}}(m,n)=\left| \mathcal{S}_m\cap  \mathcal{S}_n \right|/\left| \mathcal{S}_n \right|\). $\left| \cdot \right|$ denotes the amount of points in a 3D segment. Intuitively, assuming \(\mathcal{S}_m\) is an under-segment containing a correct segment \(\mathcal{S}_n\), \(\phi_{\mathrm{geo}}^{\mathrm{ior}}(m,n)\) will be relatively large.
Based on this, we can collect all segments contained in \(\mathcal{S}_m\) as \(\{\mathcal{S}_j~|~\phi_{\mathrm{geo}}^{\mathrm{ior}}(m,j) \geq 0.8 \}\). If the semantic feature variance of these contained segments exceeds a threshold $\tau_u$, it indicates that  \(\mathcal{S}_m\) is probably an under-segment containing different objects, and \(\mathcal{S}_m\) will be filtered out. We term this process as semantic-guided under-segment filtering.
To merge the left segments, we compute an overall similarity \(\phi=\phi_{\mathrm{geo}}^{\mathrm{iou}}+\phi_{\mathrm{geo}}^{\mathrm{ior}}+\phi_{\mathrm{sem}}^{{v}}+\phi_{\mathrm{sem}}^{{c}}\), and iteratively merge segments within group \(\mathcal{G}_i\). At each iteration, we merge highly similar segments satisfying \(\phi(m,n)\geq\theta_i\). However, simply doing so struggles to merge partially observed segments or over-segments sharing low spatial similarity. To this end, we linearly decay \(\theta_i\) at each step inspired by \cite{hovsg,sai3d}.

\subsection{Instance Feature Aggregation} \label{sec:feature_aggregate}
After obtaining the instance map \(\mathcal{M}\), each 3D instance \(\mathcal{O}_i\) in \(\mathcal{M}\) is associated with multiple segment features \(\mathcal{F}_i=\{ \mathbf{f}_{i,j}^v~|~\mathbf{f}_{i,j}^v \in \mathcal{O}_i\}\) based on 2D-3D relations (for simplicity, we omit caption features here). 
%
To aggregate these features, previous methods either perform averaging \cite{ovir3d,conceptgraph} or select the dominant feature via clustering \cite{hovsg, openmask3d}. However, they overlook the distinction between different instance features.
Hence, we propose a weighted average method to fuse an instance's features for an optimal feature both representative and distinctive, as shown in Fig. \ref{fig:overview} (b).
Specifically, taking the visual modality for illustration, we average \(\mathcal{F}_i\) to a central feature \(\bar{\mathbf{f}}_i^v\) for each instance, and the neighboring instances of \(\mathcal{O}_i\) are then defined by \(\mathcal{N}_{i} = \{ \mathcal{O}_k~|~\mathrm{cos}\left(\bar{\mathbf{f}}_{i}^{v},\bar{\mathbf{f}}_k^v\right) \geq \tau_d\} \).
Based on this, we aggregate the features \(\mathcal{F}_i\) into an optimal \({\mathbf{f}_{i}^{v}}^*\) via assigning a dynamic fusion weight \(a_{i,j}^v\) to each \(\mathbf{f}_{i,j}^{v}\) in \(\mathcal{F}_i\):
\begin{equation}
\label{rd}
    a_{i,j}^{v}=\mathrm{cos}\left( \mathbf{f}_{i,j}^{v},\bar{\mathbf{f}}_{i}^{v} \right) -\sum_{O_k\in \mathcal{N} _i}{\mathrm{cos}\left( \mathbf{f}_{i,j}^{v},\bar{\mathbf{f}}_{k}^{v} \right)},
\end{equation}
where \(\mathrm{cos}(\cdot)\) denotes cosine similarity, and \(a_{i,j}^{v}\) is normalized via softmax. Intuitively, a feature gets a larger weight if it is closer to its own cluster center and farther from neighboring instances. The caption feature \({\mathbf{f}_{i}^{c}}^*\) can be formulated by replacing \(\mathbf{f}_{i,j}^{v}\) with \(\mathbf{f}_{i,j}^{c}\) in the above process. The final instance feature \(\mathbf{f}_i^*\) is the average of \({\mathbf{f}_{i}^{v}}^*\) and \({\mathbf{f}_{i}^{c}}^*\).

\begin{figure}[t]
    \centering
    \includegraphics[width=0.45\textwidth]{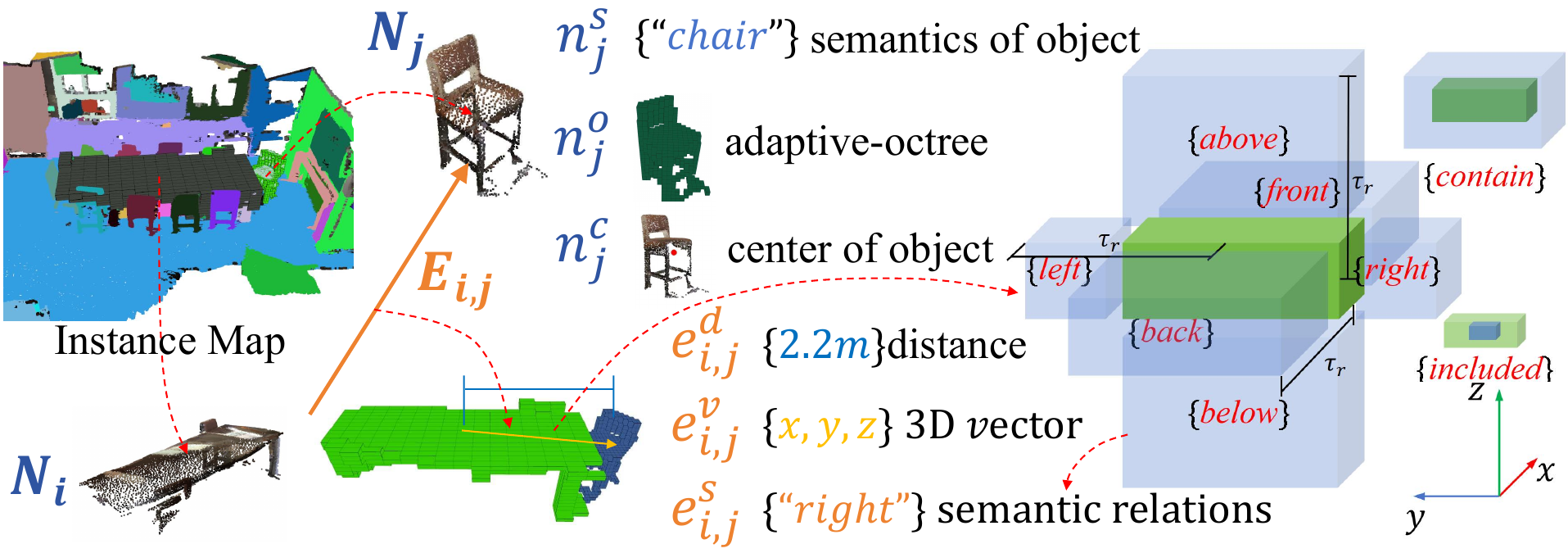} 
    \caption{Illustration of the nodes and edges in Octree-Graph.}
    \label{Figure:graph}
    \vspace{-2mm}
\end{figure}

\subsection{Octree-Graph Construction and Applications}
\label{sec:octree-graph}
To efficiently and accurately represent a scene, we design a hybrid structure, termed Octree-Graph. This structure utilizes a graph as the high-level architecture to organize objects and their spatial relations. Furthermore, we propose an adaptive-octree to depict the occupancy information of each object, which acts as a node of the Octree-Graph.


\noindent\textbf{Graph Construction.}
An Octree-Graph can be defined as \(G\) with nodes \(\mathbf{N_*}\) and edges \(\mathbf{E_*}\).
The node $\mathbf{N}_i$ consists of correlated semantics $n_i^s$ (\eg, captions and features), center $n_i^c$, and adaptive-octree $n_i^o$. While the edge $\mathbf{E}_{i,j}$ comprises the semantic relation $e_{i,j}^s$, spatial distance $e_{i,j}^d$ and the 3D vector $\mathbf{e}_{i,j}^v$ between node $i$ and node $j$.
Notably, the semantic relations between nodes are aligned with the world coordinate system of the corresponding point cloud. As shown in Fig. \ref{Figure:graph}, the semantic relation \(e_{i,j}^s\) between node \(\mathbf{N}_i\) and \(\mathbf{N}_j\) is characterized as ``right''.

\noindent\textbf{Adaptive-Octree Construction.}
The classical octree \cite{octomap} is a tree-based structure capable of efficiently representing a 3D space with much less storage requirements than point clouds. During octree construction, the root node is defined by the minimal bounding box containing the point cloud. This box, centered at \(c \in \mathbb{R}^3\) with a side length of \(d\), is divided into eight sub-regions of side length \(d/2\) using axis-aligned planes. Each sub-region serves as a child node, and the process continues recursively for each node until the desired octree depth \(L_\mathrm{max}\) is reached or no point clouds are present within the node. 
We recommend referring to \cite{octomap} for more details about the octree.

\begin{figure}[t]
    \centering
    \includegraphics[width=0.45\textwidth]{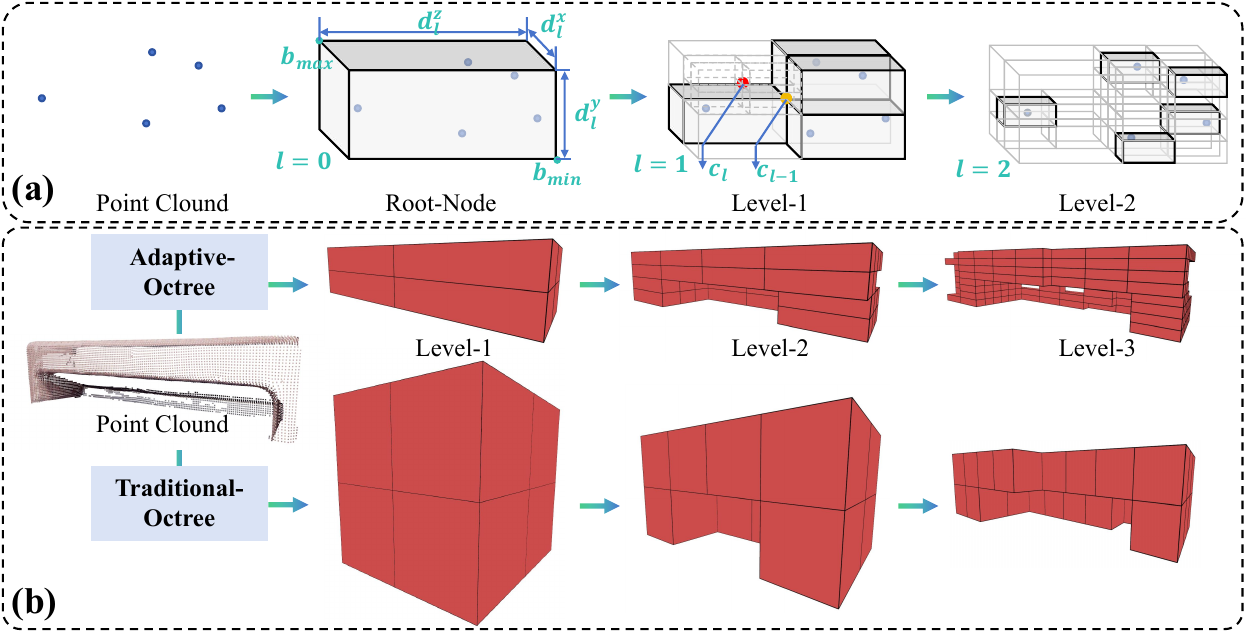} 
    \caption{Illustration of the construction of the adaptive-octree. The above displays the process, and the below shows an example.}
    \label{Figure:adaoct}
    \vspace{-2mm}
\end{figure}

However, the traditional octree is proposed to represent an entire 3D space, which uses cubic voxels as units to depict occupancy details. This leads to dilemmas of redundant representation when depicting an object, \eg, an object with a large aspect ratio requires a very deep octree to approximate its shape.
To this end, we propose the adaptive-octree with varying voxels that adaptively adjust their sizes and shapes according to the object's shape. As shown in Fig. \ref{Figure:adaoct}, an adaptive-octree is constructed from an instance-level point cloud \(P\). The size of each node in this adaptive-octree can be computed as follows:
\begin{equation}
    \mathbf{d}_l=\left(\mathbf{b}_{\mathrm{max}}-\mathbf{b}_{\mathrm{min}}\right)/ {2^l},
\end{equation}
where $\mathbf{b}_{\mathrm{max}}$ and $\mathbf{b}_{\mathrm{min}}$ are the coordinates of the lower left corner and the upper right corner of \(P\)'s bounding box. \(l\in \left \{ 1,2,\cdots ,L_\mathrm{max}  \right \} \) denotes the depth of the adaptive-octree. As shown in Fig. \ref{Figure:adaoct}, the center \(\mathbf{c}_l \in \mathbb{R}^3 \) of the \( l \)-th layer node can be determined by the center \(\mathbf{c}_{l-1} \) of the parent node and the edge length \(\mathbf{d}_l \) of the current node. The adaptive-octree can be quickly constructed from the point cloud.

\noindent\textbf{Octree-Graph Applications.}
\label{sec:og_application}
Based on the Octree-Graph, we offer object retrieval and path planning functionalities which are critical for embodied agents. For object retrieval, two types of queries are supported, \emph{i.e.,}  \(\mathrm{Query} \left(\mathit{target}\right)\) and \(\mathrm{Query} \left(\mathit{reference},\mathit{relation},\mathit{target}\right)\). The former allows for directly locating an object by comparing the similarity between queries and stored semantics. The latter supports complex queries by sequentially locating the reference object, the edge that matches the described relation, and finally the target. For more complex queries, we leverage the reasoning capabilities of LLMs to decompose the task and flexibly call two types of functions to achieve the goal.

In path planning tasks, querying occupancy information is fundamental. The proposed Octree-Graph supports such queries, enabling us to easily implement path planning algorithms like classical \(A^*\) \cite{astar} and the recent \cite{JPS}.

%% file: sec/4_experiment.tex
\begin{table*}[t]
    \centering
    \begin{tabular}{@{}lrrrrrrr@{}}
    \toprule
     &  & \multicolumn{3}{c}{\textbf{Replica}} & \multicolumn{3}{c}{\textbf{ScanNet}} \\ \cmidrule(l){3-8} 
    \textbf{Method} & \textbf{CLIP Backbone} & \textbf{mIoU}$\uparrow$ & \textbf{F-mIoU}$\uparrow$ & \textbf{mAcc}$\uparrow$ & \textbf{mIoU}$\uparrow$ & \textbf{F-mIoU}$\uparrow$ & \textbf{mAcc}$\uparrow$ \\ \midrule
    \multirow{2}{*}{ConceptFusion \cite{conceptfusion}} & OVSeg & 0.10 & 0.21 & 0.16 & 0.08 & 0.11 & 0.15 \\
     & Vit-H-14 & 0.10 & 0.18 & 0.17 & 0.11 & 0.12 & 0.21 \\ \midrule
    \multirow{2}{*}{ConceptGraph \cite{conceptgraph}} & OVSeg & 0.13 & 0.27 & 0.21 & 0.15 & 0.18 & 0.23 \\
     & Vit-H-14 & 0.18 & 0.23 & 0.30 & 0.16 & 0.20 & 0.28 \\ \midrule
    \multirow{2}{*}{HOV-SG \cite{hovsg}} & OVSeg & 0.144 & 0.255 & 0.212 & 0.214 & 0.258 & 0.420 \\
     & Vit-H-14 & 0.231 & 0.386 & 0.304 & 0.222 & 0.303 & 0.431 \\ \midrule
    \multirow{2}{*}{\textbf{Ours}} & OVSeg & \textbf{0.320} & \textbf{0.553} & \textbf{0.414} & \textbf{0.393} & \textbf{0.508} & \textbf{0.601} \\
     & Vit-H-14 & 0.263 & 0.479 & 0.387 & 0.356 & 0.477 & 0.574 \\ \bottomrule
    \end{tabular}
    \caption{Zero-shot 3D semantic segmentation results on Replica and ScanNet benchmark.}
    \label{tab:exp-semantic-segment}
    \vspace{-0.4mm}
\end{table*}

\begin{table}[t]
    \centering
    \begin{tabular}{@{}lrrr@{}}
    \toprule
    \textbf{Method} & \multicolumn{1}{c}{\textbf{AP}$\uparrow$} & \multicolumn{1}{c}{\textbf{AP50}$\uparrow$} & \multicolumn{1}{c}{\textbf{AP25}$\uparrow$} \\ \midrule
    \textit{sup. mask + sup. semantic} &  &  &   \\
    Mask3D \cite{mask3d} & 26.9 & 36.2 & 41.4 \\ \midrule
    \textit{sup. mask + z.s. semantic} &  &  &   \\
    Open3DIS \cite{open3dis} & \textbf{23.7} & 29.4 & 32.8  \\
    Open3DIS \cite{open3dis} (3D only) & 18.6 & 23.1 & 27.3  \\
    OpenMask3D \cite{openmask3d} (Mask3D) & 15.4 & 19.9 & 23.1  \\  
    \textbf{Ours} (Mask3D) & 23.2 & \textbf{30.3} & \textbf{33.3}  \\ \midrule
    \textit{z.s. mask + z.s. semantic} &  &  &   \\
    OVIR-3D \cite{ovir3d} & 9.3 & 18.7 & 25.0  \\
    SAM3D \cite{sam3d} & 9.8 & 15.2 & 20.7  \\
    SAI3D \cite{sai3d} & 12.7 & 18.8 & 24.1  \\
    Mask-Clustering \cite{maskclustering} & 12.0 & 23.3 & 30.1 \\
    \textbf{Ours} & \textbf{14.3} & \textbf{25.8} & \textbf{33.6}\\ \bottomrule
    \end{tabular}
    \caption{3D instance segmentation results on ScanNet200. \(sup.\) means supervised training, \(z.s.\) denotes the zero-shot setting.}
    \label{tab:exp-instance-segment}
\end{table}

\begin{table}[t]
\centering
\setlength{\tabcolsep}{1.5mm}{
\begin{tabular}{@{}llll@{}}
\toprule
\textbf{Method} & \begin{tabular}[c]{@{}l@{}}\textbf{Retrieval}\\ \textbf{Algorithm}\end{tabular} & \begin{tabular}[c]{@{}l@{}}\textbf{Acc$\uparrow$}\\ \textbf{@0.1}\end{tabular} & \begin{tabular}[c]{@{}l@{}}\textbf{Acc$\uparrow$}\\ \textbf{@0.25}\end{tabular} \\ \midrule
ConceptGraphs \cite{conceptgraph} & \textit{Deductive} \cite{bbq}     & 0.15 & 0.08 \\ 
Open-Fusion \cite{openfusion}  & CLIP \cite{CLIP}          & 0.13 & 0.02 \\
BBQ \cite{bbq}           & CLIP \cite{CLIP}         & 0.10 & 0.06 \\
BBQ \cite{bbq}           & \textit{Deductive} \cite{bbq}    & 0.23 & 0.18 \\
\textbf{Ours}          &   Octree-Graph+LLM       & \textbf{0.26} & \textbf{0.23} \\ \bottomrule
\end{tabular}
}
\caption{Text-based object retrieval results on the Sr3D dataset.}
\label{tab:exp-retrieval}
\vspace{-2mm}
\end{table}

\begin{table}[t]
\centering
\setlength{\tabcolsep}{0.95mm}{
\begin{tabular}{@{}llll@{}}
\toprule
\textbf{Method} & \textbf{SR($s$=1.0m)} & \textbf{SR($s$=0.5m)} & \textbf{SR($s$=0.25m)} \\ \midrule
HOV-SG \cite{hovsg}    &55.25     & 46.75 & 32.16 \\
\textbf{Ours}          & \textbf{97.88}   & \textbf{96.88} & \textbf{96.38} \\ \bottomrule
\end{tabular}
}


\caption{Path planning results on HM3DSem. SR denotes success rate (\%). $s$ is the threshold within which the distance between the navigation endpoint and the destination is considered successful.}
\label{tab:exp-path}
\vspace{-2mm}
\end{table}

\section{Experiment}
\label{sec:exp}
To validate the versatility and effectiveness of our method, we carry out extensive experiments, including semantic segmentation, instance segmentation, text-based object retrieval, and path planning. We compare our method with different SOTA methods in these tasks, and conduct comprehensive ablation studies to investigate several key components, demonstrating the effectiveness of our designs.

\subsection{Implementation Details}
We use CropFormer \cite{Cropformer} as the 2D proposal generator following \cite{maskclustering}. To extract visual features, we test two commonly used VLMs, \emph{i.e.,} CLIP ViT-H \cite{CLIP} and OVSeg ViT-L \cite{OVSeg}. We adopt TAP \cite{Tap} for generating the mask caption.
Additionally, we filtered out masks with pixels less than \(25\) and segments with points less than \(50\). The similarity threshold \(\tau_d\) is empirically set to \(0.7\). The group split interval $I$, the under-segment filtering threshold \(\tau_u\), and the decay parameter $\theta_i$ are set to 200, 0.02, and 0.8 through hyper-parameter experiments. Considering the dimensions of indoor objects, we set the maximum depth \(L_\mathrm{max}\) of the adaptive-octree to 4.

\subsection{Dataset and Evaluation Metrics}

\textbf{Dataset.} 
For zero-shot 3D semantic segmentation, we evaluate our method on common scenes following \cite{conceptfusion, conceptgraph, hovsg}, \textit{i.e.},  $8$ scenes from Replica \cite{replica} dataset and $5$ scenes from ScanNet \cite{scannet}. For zero-shot 3D instance segmentation, we assess our method on the widely-used ScanNet200 \cite{scannet200} benchmark, including a validation set of $312$ scenes with $200$ categories. For text-based object retrieval, we test our method on Sr3D \cite{sr3d} dataset, and follow the experiment setting of BBQ \cite{bbq} that subsampled 526 free-form queries from 8 scenes. For the path planning task, we employ the HM3DSem \cite{hm3dsem} dataset used in HOV-SG \cite{hovsg}, where 8 scenes are selected for evaluation. Moreover, we also conduct real-world experiments to validate our effectiveness.

\noindent \textbf{Evaluation Metrics.}
Following the mainstream evaluation metrics \cite{hovsg}, we assess 3D semantic segmentation results via commonly used mean IoU (mIoU), frequency-weighted mean IoU (F-mIoU), and mean Accuracy (mAcc). For 3D instance segmentation, we report the standard Average Precision (AP) at IoU thresholds 25$\%$ and 50$\%$, along with the mean of AP from 50$\%$ to 95$\%$ at 5$\%$ interval. 
For text-based object retrieval, we follow BBQ \cite{bbq}, using Acc@0.1 and Acc@0.25 as evaluation metrics where retrieval is treated as a true positive if the IoU between the predicted object's bounding box and the ground-truth bounding box surpasses 0.1 and 0.25, respectively. For the path planning task, we randomly select positions in the empty areas of a scene as the starting point and destination. When the endpoint of navigation is within a threshold $s$ (\emph{i.e.}, 1m, 0.5m, and 0.25m) from the destination, the path planning is considered successful.

Besides, to quantify spatial representation accuracy, we introduce the Effective Occupancy Ratio (EOR) as a metric. The occupancy range \(O_\mathrm{pc}\) of a point cloud is determined by expanding this point cloud with a dilation \(\bigtriangleup r=0.005\), and the occupancy range of the octree is denoted as \(O_\mathrm{oct}\). Then, the EOR is calculated as \(\mathrm{EOR} =\frac{O_\mathrm{oct}\cap O_\mathrm{pc} }{O_\mathrm{oct}} \). We denote the mean EOR for all objects in a scene as mEOR.

\subsection{Quantitative Comparison} 
\textbf{3D Semantic Segmentation.}
Tab. \ref{tab:exp-semantic-segment} reports the numerical results for zero-shot 3D semantic segmentation on Replica and ScanNet datasets. In this experiment, we compare the results generated by our CGSM and IFA with other works. It can be seen that our method significantly outperforms existing methods across all metrics on both datasets, demonstrating the effectiveness of the proposed CGSM and IFA.
Compared to the existing SoTA 3D scene graph, HOV-SG \cite{hovsg}, we achieve +$\textbf{8.9\%}$ mIoU and +$\textbf{11.0\%}$ mAcc on the Replica dataset. Similarly, we present a +$\textbf{17.1\%}$ mIoU and a +$\textbf{17.0\%}$ mAcc on ScanNet with the same settings.

\noindent\textbf{3D Instance Segmentation.}
The quantitative results for 3D instance segmentation are shown in Tab. \ref{tab:exp-instance-segment}. We follow \cite{maskclustering} to categorize all methods into 3 groups based on whether the proposal generation and semantic prediction are trained.
Under the fully zero-shot setting, our method surpasses the previous most advanced method with gains of $\textbf{2.3\%}$, $\textbf{2.5\%}$, and $\textbf{3.5\%}$ in AP, AP25 and AP50, respectively.
These results further demonstrate the effectiveness of the proposed CGSM and IFA. 
When using supervised 3D models for proposal generation, our method significantly outperforms OpenMask3D \cite{openmask3d} and the Open3DIS \cite{open3dis} variant with only the 3D proposals, validating the superiority of our feature aggregation method IFA. Besides, our method achieves comparable results with the corresponding SOTA method Open3DIS \cite{open3dis}, which specially designs a combination of 2D and 3D proposals.

\noindent\textbf{Text-based Object Retrieval.} Tab. \ref{tab:exp-retrieval} presents the comparison results of text-based object retrieval on Sr3D \cite{sr3d} dataset. Our method outperforms the SOTA method BBQ \cite{bbq} by $\textbf{3.0\%}$ and $\textbf{5.0\%}$ in terms of Acc@0.1 and Acc@0.25, respectively. This experiment veriffes the quality of our con-structed
graph. We attribute the performance gain to our accurate semantic object segmentation and the rich relations stored in the Octree-Graph. 

\noindent\textbf{Path Planning.} For each sense in the HM3DSem \cite{hm3dsem} dataset, we randomly select 100 pairs of starting points and destinations in navigable areas. HOV-SG \cite{hovsg} can be directly used for path planning, thus it is evaluated and compared with our method in this task. Tab. \ref{tab:exp-path} shows the results, from which we can see that our method significantly surpasses HOV-SG, especially when the threshold $s$ is small. This is because HOV-SG relies on Voronoi graph \cite{voronoi} for path planning, where the waypoints and paths are pre-calculated, making it improper for precise navigation. In contrast, our Octree-Graph supports navigation to any empty area, unless the destination is mistakenly occupied by the adaptive-octree.  

\begin{figure*}[t]
    \centering
    \includegraphics[width=0.99\textwidth]{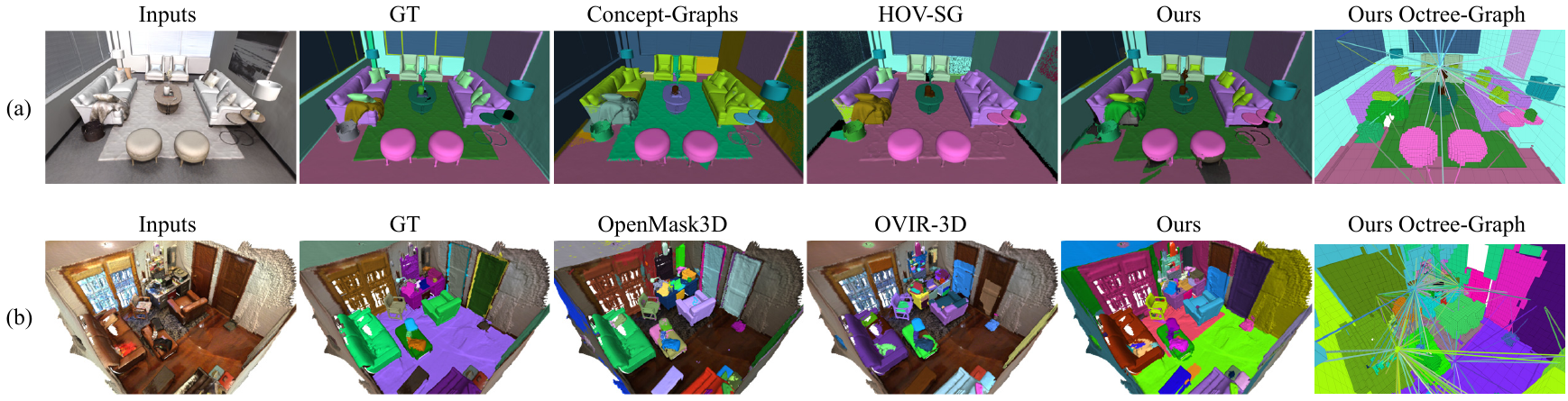} 
    \caption{Visual comparisons. (a) Semantic segmentation results on Replica. (b) Instance segmentation results on ScanNet200.}
    \label{fig:visualize}
\end{figure*}

\begin{table}[t]
    \centering
    \begin{tabular}{@{}llll@{}}
    \toprule
    \textbf{Merging Strategy} & \textbf{mIoU}$\uparrow$ & \textbf{F-mIoU}$\uparrow$ & \textbf{mAcc}$\uparrow$ \\ \midrule
    Frame-wise        & 0.323 & 0.439 & 0.519  \\
    Global-wise           & 0.286 & 0.414 & 0.476  \\
    Ours (CGSM) $I$=400       & 0.338  & 0.463 & 0.555  \\
    \textbf{Ours} (CGSM) $I$=200       & \textbf{0.356}  & \textbf{0.477} & \textbf{0.574} \\
    Ours (CGSM) $I$=100       & 0.344  & 0.462 & 0.571     \\ \bottomrule
    \end{tabular}
    \caption{Ablation study on the segment merging strategy and different temporal intervals for group partitioning of our CGSM.}
    \label{tab:ablation-merging}
    \vspace{-1mm}
\end{table}

\begin{table}[t]
    \centering
    \setlength{\tabcolsep}{1mm}{
    \begin{tabular}{@{}ccccc@{}}
    \toprule
    \begin{tabular}[c]{@{}c@{}}\textbf{under-segment}\\ \textbf{filtering}\end{tabular} & \begin{tabular}[c]{@{}c@{}}\textbf{threshold} \\ \textbf{decay}\end{tabular} & \textbf{mIoU}$\uparrow$ & \textbf{F-IoU}$\uparrow$ & \textbf{mAcc}$\uparrow$ \\ \midrule
    \ding{55} & \ding{55}     & 0.337 & 0.460 & 0.547 \\
    \checkmark & \ding{55}    & 0.346 & 0.471 & 0.557 \\
    \checkmark & \checkmark   & \textbf{0.356} & \textbf{0.477} & \textbf{0.574}  \\ \bottomrule
    \end{tabular}
    }
    \caption{Ablations study on the strategies for segment merging.}
    \label{tab:merging-strategy}
    \vspace{-1mm}
\end{table}

\begin{table}[t]
\centering
    \begin{tabular}{@{}llll@{}}
    \toprule
    \textbf{Aggregation Strategy} & \textbf{mIoU}$\uparrow$ & \textbf{F-mIoU}$\uparrow$ & \textbf{mAcc}$\uparrow$ \\ \midrule
    Average              &  0.338  &  0.453  &  0.563    \\
    Top-5                &  0.342  &  0.470  &  0.570    \\
    DBSCAN               &  0.345  &  0.459  &  0.566    \\
    \textbf{Ours} (IFA)              &  \textbf{0.356 } & \textbf{ 0.477}  &  \textbf{0.574}      \\ \bottomrule
    \end{tabular}
    \caption{Ablation study on various feature aggregation strategies.}
    \label{tab:ablation-aggregation}
    \vspace{-1mm}
\end{table}

\begin{table}[t]
    \centering
    \setlength{\tabcolsep}{0.9mm}{
    \begin{tabular}{@{}lllll@{}}
    \toprule
                    & \multicolumn{2}{c}{\textbf{Replica}}  & \multicolumn{2}{c}{\textbf{ScanNet}}  \\ \cmidrule(l){2-5} 
    \textbf{Method}  & \textbf{Storage}$\downarrow$ & \textbf{mEOR$\uparrow$}  & \textbf{Storage$\downarrow$} & \textbf{mEOR$\uparrow$}             \\ \midrule
    Point cloud     & 18.5MB    &                  & 6.4MB     &                  \\
    Octree          & 17.6KB    & 0.0057          & 41.1KB    & 0.0041          \\
    Adaptive-Octree & 29.8KB    & \textbf{0.0108} & 69.3KB    & \textbf{0.0070} \\ \bottomrule
    \end{tabular}
    }
    \caption{Ablation study on the efficiency of the adaptive-octree.}
    \label{tab:octree-table}
    \vspace{-1mm}
\end{table}

\begin{table}[t]
    \centering
    \setlength{\tabcolsep}{0.9mm}{
    \begin{tabular}{@{}lllll@{}}
    \toprule
        \textbf{Method}             & \textbf{Structure} & \textbf{Storage}$\downarrow$ & \textbf{time}$\downarrow$ \\ \midrule
        A*                          & Octree-Graph       &  268.41Kb         & 0.032s         \\
        A*                          & Point Cloud        &  71.16Mb          & 2.154s         \\ \midrule
        Jump Point Search           & Octree-Graph       &  268.41Kb         & 0.081s         \\ 
        Jump Point Search           & Point Cloud        &  71.16Mb          & 2.153s         \\ \bottomrule
        \end{tabular}
    }
    \caption{Ablation study on path planning efficiency.}
    \label{tab:path_planning_com}
\end{table}


\subsection{Ablation Studies}
We analyze the impact of our key designs via zero-shot semantic segmentation experiments on ScanNet.

\noindent\textbf{Effect of Group-Wise Split.}
We compare the proposed Chronological Group-wise Segment Merging (CGSM) with the vanilla frame-wise and global-wise merging strategies. 
As shown in Tab. \ref{tab:ablation-merging}, we achieve notable gains compared to both methods, with +$\textbf{3.3\%}$ mIoU over frame-wise merging and  +$\textbf{7.0\%}$ mIoU over global-wise merging.
We also analyze the impact of hyper-parameter \( I \), and the results in Rows 3-5 show that our method exhibits robustness to \( I \) ranging from 100 to 400.

\noindent \textbf{Analysis of Designs on Segment Merging.}
Tab. \ref{tab:merging-strategy} presents the results with two key components for merging a single group. Row $0$ serves as a fixed-threshold group-wise merging baseline with no extra design. Row $1$ is equipped with our semantic-guided under-segment filtering, and achieves +$\textbf{0.9}\%$ mIoU and +$\textbf{1.0}\%$ mAcc. Row $2$ further incorporates the threshold decay strategy, resulting in an additional $\textbf{1.0}\%$ mIoU and $\textbf{1.7}\%$ mAcc gains. These validate the effectiveness of the two strategies during each group-wise merging.

\noindent \textbf{Effect of Instance Feature Aggregation.}
In Tab. \ref{tab:ablation-aggregation}, we compare the proposed Instance Feature Aggregation (IFA) method with several commonly used methods. For fairness, all methods use the same features as our IFA. Row $1$ simply averages features across all views, yielding unsatisfactory results. Row $2$ and Row $3$ leverage Top-5 criterion \cite{mask3d, openmask3d} and DBSCAN algorithm \cite{hovsg} to select the predominant feature, achieving gains of +$\textbf{0.4}\%$ and +$\textbf{0.7}\%$ mIoU, respectively. By contrast, our IFA achieves an improvement of $\textbf{1.8}\%$ mIoU over Row $1$.  

\noindent\textbf{Adaptive-Octree Efficiency.}
Based on the results of instance generation in Replica and ScanNet, Tab. \ref{tab:octree-table} provides a comparison of different spatial representations with respect to storage space and the accuracy of occupancy. For the same scene, octree and our adaptive-octree consume two orders of magnitude less storage compared to point clouds. Our adaptive-octree requires a bit more storage than the traditional octree due to its additional record of bounding boxes for each object. However, at the same depth, the adaptive-octree exhibits a much higher mEOR compared to the octree. This means that the space described by the adaptive-octree is more closely aligned with the target regions. In summary, the adaptive-octree requires much less storage space than point clouds and provides more accurate occupancy information than a traditional octree.

\noindent\textbf{Path Planning Efficiency based on Octree-Graph.} To further verify the efficiency of our method, we conduct path planning experiments using A* \cite{astar} and Jump Point Search \cite{JPS} algorithms on the HM3DSem \cite{hm3dsem} dataset, where the point cloud representation and our Octree-Graph are compared. It can be seen from Tab. \ref{tab:path_planning_com} that Octree-Graph uses much less storage space and spends much less time. This is crucial for real-world deployment.


\subsection{Qualitative Analysis}
\begin{figure}[t]
    \centering
    \includegraphics[width=0.45\textwidth]{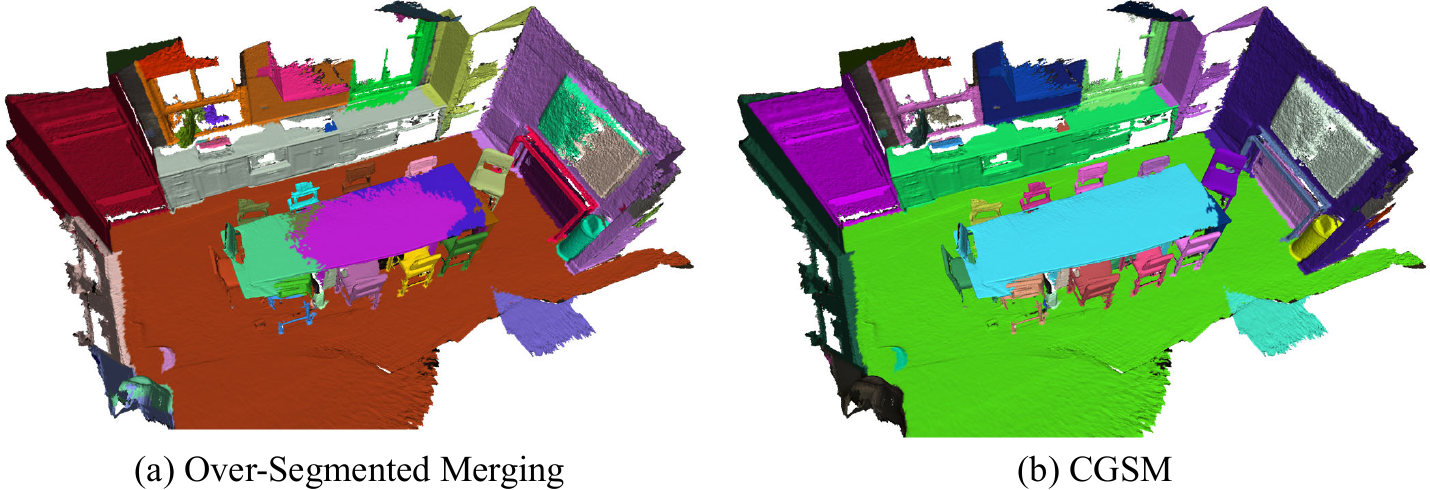} 
    \caption{Segment merging comparison.}
    \label{fig:seg_com}
    \vspace{-2mm}
\end{figure}

\noindent\textbf{Visualization Results.} Fig. \ref{fig:visualize} visualizes the results of semantic segmentation and instance segmentation, respectively. We can see that our method exhibits more accurate object semantics and fewer incorrect segments than comparison methods. Fig. \ref{fig:seg_com} demonstrates the segment merging results of our CGSM and its baseline (\emph{i.e.,} frame-wise sequential merging), where CGSM correctly resolves the over-segmented long table without introducing excessive merges between different objects.

\begin{figure}[t]
    \centering
    \includegraphics[width=0.45\textwidth]{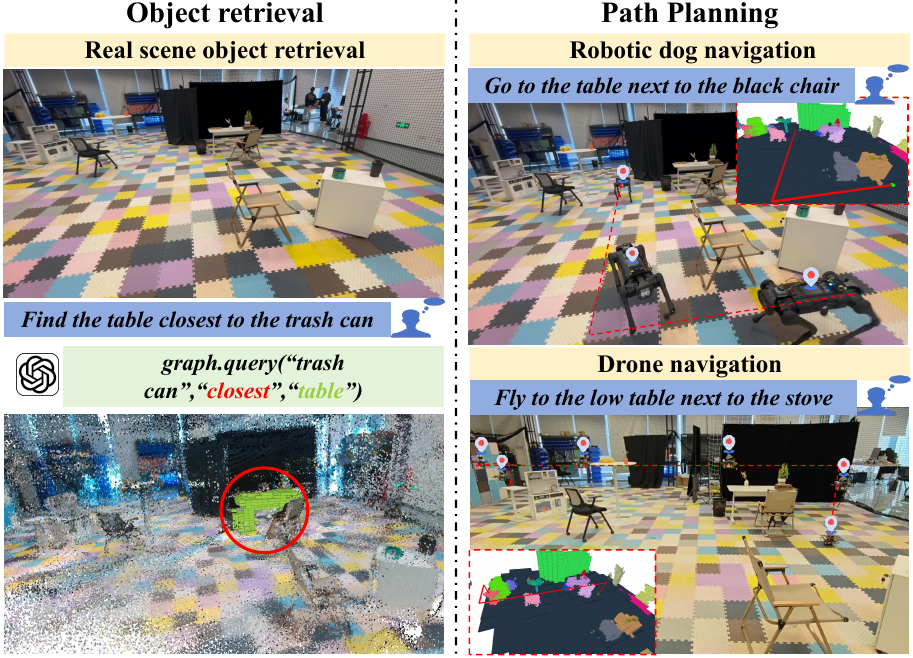} 
    \caption{Visualization of the real-world experiment. The left column shows a real scene and the reconstructed colored point cloud with the retrieval target highlighted by our adaptive-octree. The right column presents path planning using a robotic dog and a drone based on our Octree-Graph.}
    \label{Figure:exp_query}
\end{figure}

\noindent\textbf{Object Retrieval and Path Planning.}
We also conduct real-world experiments to further validate the effectiveness of our method. Fig. \ref{Figure:exp_query} presents the results where a real scene is set up and scanned by an Intel Realsense D435i camera. Then we reconstruct the colored point cloud and establish the Octree-Graph. Based on these, we deploy our method on a robotic dog and a drone with NVIDIA Orin NX as onboard computers. As shown in Fig. \ref{Figure:exp_query}, robots can accurately find the target and successfully navigate to it relying on our Octree-Graph. The dynamic process of this experiment can be found in the supplementary video demo.

%% file: sec/5_conclusion.tex
\section{Conclusion}
\label{sec:conclusion}

In this paper, we propose Octree-Graph, a novel scene representation for open-vocabulary 3D scene understanding. Specifically, an adaptive-octree structure is devised to characterize the occupancy of an object, which acts as the node of the Octree-Graph. The edges describe rich relations among objects for spatial reasoning. For Octree-Graph construction, we also develop a training-free pipeline to conduct semantic object segmentation, where a Chronological Group-wise Segment Merging (CGSM) strategy is designed to alleviate inaccurate segment proposals, and an Instance Feature Aggregation (IFA) method is devised to get a semantic feature both representative and distinctive. Extensive evaluations on several tasks validate the versatility and effectiveness of our Octree-Graph. 